\documentclass[letterpaper, 10 pt, journal]{IEEEtran}
\usepackage{float}

\IEEEoverridecommandlockouts                              

\usepackage{graphics} 
\usepackage{graphicx} 
\usepackage{amssymb}  
\usepackage{subcaption}
\usepackage{makecell}
\usepackage{verbatim}
\usepackage{amsmath} 
\usepackage{todonotes}

\title{ \LARGE \bf CAOS-SLAM: Class-Agnostic Object-Aware Real-Time vSLAM for Dynamic Environments}

\author{ Gaia Borghi$^{1,2*}$, João Carlos Virgolino Soares $^1$, Angelo Bratta $^1$, Shahab Heshmati-Alamdari $^2$, Claudio Semini $^1$
\thanks{*Corresponding author: {\tt\small gaia.borghi@iit.it}\\
$^1$ Dynamic Legged Systems Lab (DLS), Istituto Italiano di Tecnologia, Genoa, Italy\\
$^2$ Section of Automation \& Control, Department
of Electronic Systems, Aalborg University, Denmark}}

\begin{document}

\maketitle

\pagestyle{empty}
\thispagestyle{empty}

\begin{abstract}

Robust visual Simultaneous Localization and Mapping (vSLAM) in dynamic environments faces challenges due to the presence of moving objects that may introduce erroneous observations, corrupt the map reconstruction and overall degrade localization accuracy. Existing vSLAM approaches often rely on predefined semantic classes or costly pixel-wise classification, limiting their robustness or applicability in real-world scenarios. This paper presents CAOS-SLAM, a dynamic vSLAM framework that extends ORB-SLAM3 with a hybrid perception pipeline to detect, track, and filter both semantically known and previously unseen moving objects in real-time. On-demand panoptic segmentation is performed for object discovery, with candidate unknown objects instantiated through depth-based clustering of unclassified regions. Between panoptic updates, persistent object tracks are updated using lightweight detection when available or temporal mask propagation based on world-frame Kalman prediction, depth consistency, and appearance similarity. The estimated object motion is then used to reject dynamic features from the SLAM pipeline. Experimental results on multiple dynamic RGB-D datasets and real-world experiments demonstrate improved localization accuracy over state-of-the-art dynamic vSLAM approaches, achieving up to 59\% lower localization error in scenes characterized by unknown dynamic objects and up to 88\% lower error in heavily occluded scenes, while achieving a median processing rate of 30 FPS on a laptop. Furthermore, the tracking module provides estimates of object trajectories and velocities, supporting potential future applications in perception, navigation, and motion-planning tasks in dynamic environments. 
Code and dataset will be released upon acceptance.

\end{abstract}


\vspace{-0.5cm}
\section{INTRODUCTION}

Visual Simultaneous Localization and Mapping (vSLAM) has become a fundamental component for autonomous robots and intelligent systems operating in unknown environments. vSLAM enables robots and autonomous devices to estimate their pose inside an unknown environment, while progressively reconstructing the surroundings only relying on visual data, without external sensors.

Despite significant progress in recent years, vSLAM systems still face notable challenges when deployed in real-world scenarios. The presence of moving objects reduces the availability of reliable static features, which might lead to incorrect data association, localization drift and degraded tracking performance \cite{Qin2024rsoslam}. Many recent approaches use object detection or instance segmentation models to identify and filter dynamic features \cite{Zhang2025ngdslam} \cite{10475327} \cite{10947207}. While these methods can achieve real-time or close to real-time performance, they inherently depend on predefined semantic classes. As a consequence, their ability to be generalized into unseen or unfamiliar scenarios is limited, especially when dynamic objects fall outside the categories on which the network was trained~\cite{9858152} \cite{Tang_2025}, \cite{11357484}.

Class-agnostic approaches aim to overcome these limitations by detecting motion without relying on semantic labels. These methods often combine deep learning with geometric cues and optical flow. These methods improve generalization to arbitrary dynamic objects, but this robustness comes with a high computational cost, which hinders their applicability in real-time systems \cite{Zhu_2024}, \cite{Wang2024usdslam}, \cite{Li2025ospslam}.

\begin{figure}[!t]
    \centering
    \includegraphics[width=1.0\linewidth, height=5.7cm]{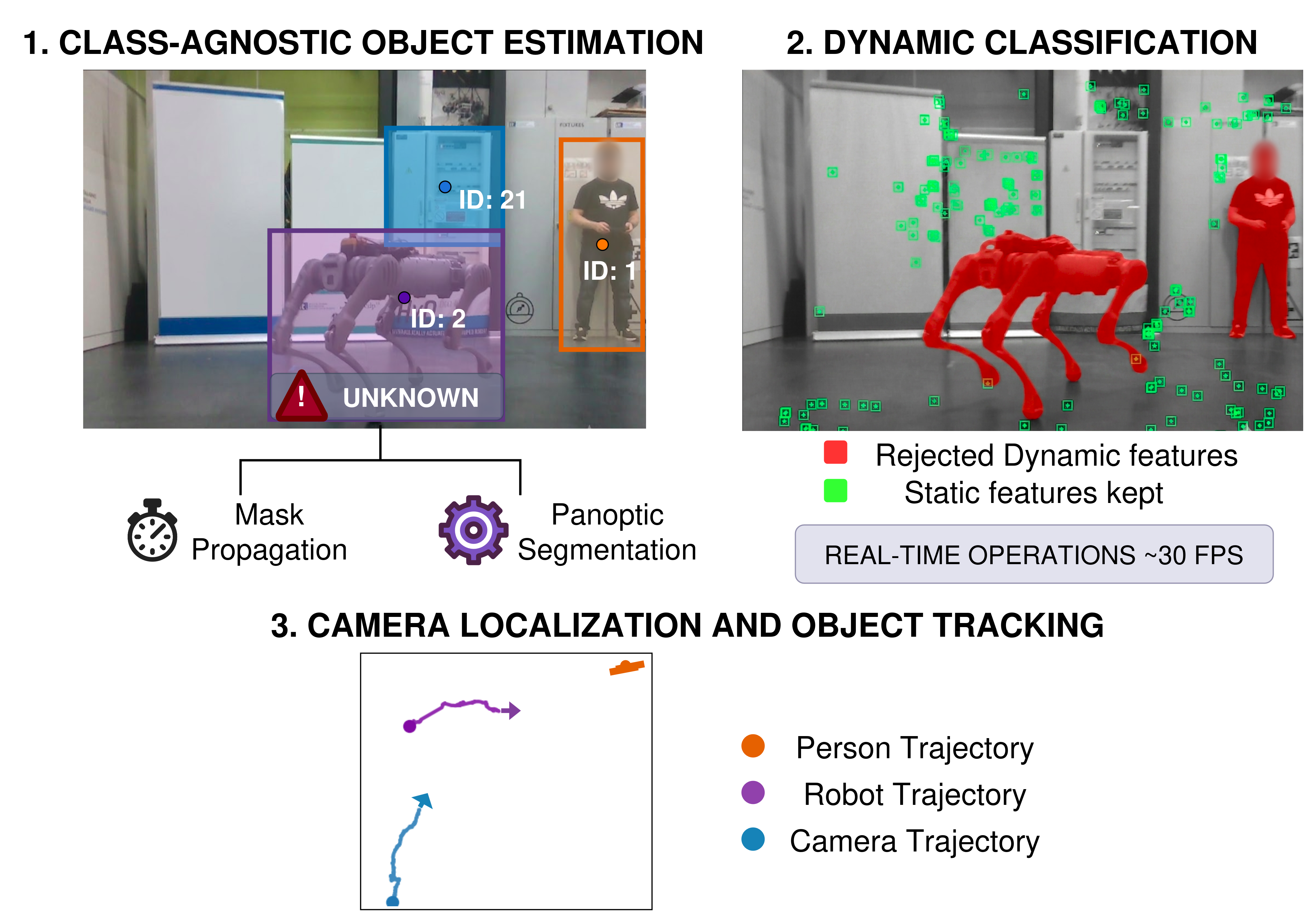}
    \caption{Object masks (known and unknown) are extracted via panoptic discovery or temporal propagation. 
    Moving objects are excluded from SLAM localization. Both camera and dynamic object trajectories are estimated.}
    \label{fig:intro_image}
\end{figure}

In this work, we address the trade-off between class-agnostic robustness and real-time performance by proposing CAOS-SLAM, a \textbf{C}lass-\textbf{A}gnostic \textbf{O}bject-aware dynamic v\textbf{S}LAM framework (Fig.~\ref{fig:intro_image}). Our approach relies on a panoptic segmentation model that provides object-level masks. To reduce dependency on predefined semantic categories, the framework processes regions that are not assigned to any known panoptic classes, using RGB-D information to instantiate candidate unknown objects. 
However, running panoptic segmentation on every frame can introduce a substantial computational overhead, making real-time operation challenging~\cite{abati2024panopticslam}. Therefore, our approach combines panoptic segmentation with a lightweight temporal mask propagation strategy. This design enables class-agnostic dynamic object detection and tracking while significantly reducing the computational cost associated with panoptic perception, allowing the complete vSLAM system to operate in real time. 
The key contributions of the proposed implementation are as follows:
\begin{itemize}
    \item \textbf{Class-agnostic dynamic reasoning:} 
    We introduce a perception framework that combines panoptic segmentation with RGB-D geometric reasoning to detect both known and unknown objects. Unclassified regions are separated into individual object instances through depth-based clustering, enabling dynamic object reasoning beyond predefined semantic classes.
    \item \textbf{Object-centric temporal propagation:} The proposed system decouples object discovery from tracking to avoid running costly panoptic inference at every frame. The system maintains an explicit per-object state and propagates object masks between panoptic observations through Kalman filter prediction combined with depth and appearance consistency, keeping object states updated while preserving real-time operation. 
    \item \textbf{Integrated dynamic object tracking:} 
 Unlike previous tracking-based methods, which are limited to predefined semantic categories \cite{9385844,10.1109/ICRA48506.2021.9561452} or rely on computationally expensive formulations \cite{11288097}, our system tracks both known and unknown objects and estimates their motion in the world frame. It runs in real time on a laptop, providing object-level information that can support other tasks such as navigation and obstacle avoidance.
    
    \item \textbf{Extensive Validation:} Experiments on multiple dynamic RGB-D datasets and real-world scenarios demonstrated improved localization accuracy and robustness in dynamic and heavily occluded scenes, including up to 59\% lower localization error for unknown dynamic objects and up to 88\% lower error under severe occlusions, compared to other SOTA methods. 
\end{itemize}

Additionally, the code and the recorded dataset will be released upon acceptance.

\section{Literature Review}
Traditional visual Simultaneous Localization and Mapping (vSLAM) systems, such as ORB SLAM3~\cite{orbslam3}, rely on the assumption of static environments. This assumption leads to severe performance degradation in real-world scenarios where moving objects confound data association, cause significant localization drift, and corrupt the 3D map. This issue becomes particularly critical in highly dynamic scenes where the number of reliable static features for pose estimation is severely limited \cite{Qin2024rsoslam, Li2025ospslam, Li2025psslam}. Consequently, research has focused on identifying, modeling and filtering dynamic entities, to ensure tracking robustness and map reliability~\cite{Qin2024rsoslam},~\cite{Zhang2025ngdslam},~\cite{Li2025ospslam}. Despite significant progress, achieving robust, universal, and real-time dynamic vSLAM remains an open problem.

A prominent direction in the vSLAM field relies on semantic information obtained from deep learning methods to identify and remove dynamic regions. Representative examples include NGD-SLAM~\cite{Zhang2025ngdslam}, DRV-SLAM~\cite{10475327}, SDT-SLAM~\cite{11357484}, CD-SLAM~\cite{10521603} and the method proposed by Su et al.~\cite{9858152}, which combine object detection or instance segmentation with geometric or motion constraints. While these approaches can achieve real-time performance and high accuracy, they are limited by the ability of the network to detect objects. Therefore, they rely on predefined object classes (e.g. humans, chairs, cars) and often fail to handle arbitrary moving objects like robots or manipulated furniture.

To overcome the limitations of semantic priors, a second line of research adopts class-agnostic motion reasoning. PS SLAM~\cite{Li2025psslam} and Panoptic SLAM~\cite{abati2024panopticslam} utilize Panoptic segmentation models and epipolar geometry to detect and filter features from known and unknown dynamic areas. However, they are highly sensitive to the panoptic model accuracy and incur significant computational overhead.
To enhance generalization, other approaches, such as OSP-SLAM~\cite{Li2025ospslam} and RSO-SLAM~\cite{Qin2024rsoslam}, combine optical flow, segmentation, and geometric motion reasoning. However, they typically require computationally demanding perception modules that limit real-time performance on onboard platforms. As a result, a practical trade-off emerges between real-time operation and real-world robustness. Methods that achieve real-time performance often do so by sacrificing universality and relying on lightweight semantic models \cite{10947207, 11357484, 9858152}, whereas more general approaches rely on computationally expensive panoptic segmentation or motion estimation \cite{Qin2024rsoslam, abati2024panopticslam, Li2025ospslam}. CAOS-SLAM addresses these limitations by combining class-agnostic object discovery with lightweight temporal tracking, aiming to achieve both the generalization capabilities required for real-world environments and the computational efficiency required for real-time vSLAM.

Efficiency on onboard platforms remains a key challenge for real-world deployment. 
Recent work reduces inference cost via temporal mask propagation \cite{Zhang2025ngdslam, Li2025ospslam, 11208982} and by decoupling semantic inference from tracking \cite{Zhang2025ngdslam, Xue2025NSDMSLAM}.
NGD-SLAM~\cite{Zhang2025ngdslam} tracks keypoints from the latest mask to the current frame with optical flow, reaching 60 FPS on a laptop CPU. NSDM-SLAM~\cite{Xue2025NSDMSLAM} decouples semantic recognition from tracking through a non-blocking buffer. As both works acknowledge, masks propagated from earlier frames can miss newly appearing objects and degrade under fast scene changes. CAOS-SLAM, instead of relying on buffered masks, invokes the panoptic model only when object discovery is required.

Another challenge arises when the scene is highly dynamic. The number of remaining static features can be insufficient for accurate localization. To improve feature matching, DRV-SLAM~\cite{10475327} adapts the number of extracted features, OSP-SLAM~\cite{Li2025ospslam} uses a deep learning network for feature extraction and matching, and PS-SLAM~\cite{Li2025psslam} introduces semantic-aware feature descriptors. Our approach is complementary: by detecting dynamic obstruction as soon as their motion becomes observable, it aims to reduce the number of dynamic artifacts in the map, so that the system can re-localize and subsequently correct drift through loop closure, once the obstruction is gone.

Beyond improving localization, dynamic object tracking provides valuable information for scene understanding and navigation. Rather than treating dynamic entities solely as outliers to be removed \cite{11288097}, recent approaches exploit object motion as an additional source of information. DynaSLAM II~\cite{9385844} proposes a tightly coupled framework in which camera poses improve object state estimations, while object poses provide additional constraints for camera localization. Similarly, DynoSAM~\cite{11288097} and DOT~\cite{10.1109/ICRA48506.2021.9561452} jointly estimate camera and object trajectories. Although these approaches still face the real-time versus robustness trade-off discussed above, the object-level information they provide can benefit other robotic tasks such as navigation and obstacle avoidance.

\section{METHODOLOGY} 
\begin{figure*}[t]
    \centering
    \includegraphics[width=1.0\textwidth]{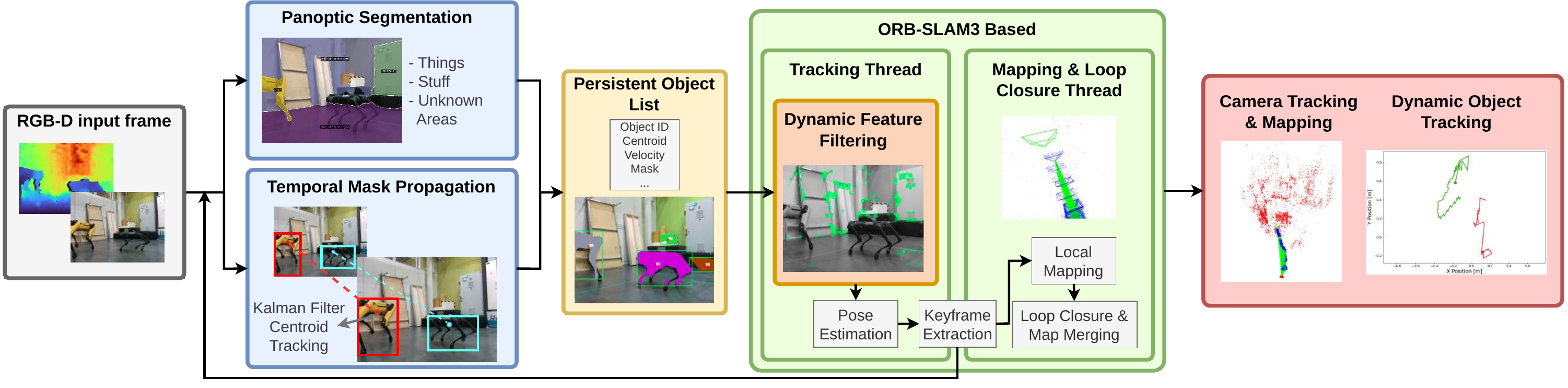}
    \caption{The proposed system is integrated into ORB-SLAM3. A panoptic perception module periodically discovers semantically known and unknown movable objects (blue block) in the scene, while a lightweight propagation module updates object masks between panoptic inferences. Object identities and motion estimates are maintained in a persistent object list (yellow block). Dynamic object masks are then used to remove dynamic ORB features from the localization and mapping framework.}
    \label{fig:system_overview}
\end{figure*}

This section presents the proposed dynamic vSLAM framework, which we integrated into the Tracking thread of ORB-SLAM3~\cite{orbslam3}. Our implementation combines panoptic segmentation, lightweight object detection, temporal mask propagation and motion-based object classification to remove dynamic features (See Fig. \ref{fig:system_overview}).

\subsection{System Overview} \label{initial_processing}
For each new RGB-D frame, ORB features are first extracted and the initial estimate of the camera pose with respect to the world reference frame, $\hat{T_{WC}}$, is obtained using the ORB-SLAM3 internal camera motion model. The frame is analyzed to detect objects in the scene and determine their motion state. 

Detected objects are saved in a persistent object database, to ensure consistency in object tracking across consecutive frames:
\begin{equation}
    object_i^k = [ID_i, Class_i, M_{i}^k, KF_{i}^k, {c}_{k_i}^{cam}, isDyn_i^k]
\end{equation}

\noindent where $ID_i$ denotes the unique object identifier, $Class_i$ its semantic category when available, $M_i^k$ the object mask, $KF_i^k$ the associated Kalman filter model, ${c}_{k_i}$ the object centroid in the camera frame and $isDyn_i^k$ its dynamic state at time $k$.\\

To preserve real-time performance while maintaining a class-agnostic strategy, the proposed implementation provides two distinct modules for object tracking:
\begin{itemize}
    \item Panoptic Module: executed periodically, to detect and initialize scene objects
    \item Propagation Module: executed on the intermediate frames, to propagate and update object states.
\end{itemize}
The tracked object states identify dynamic regions and remove the corresponding ORB features before localization and mapping, leaving only static features for the SLAM estimation.

\subsection{Panoptic Module}

The Panoptic module is executed only if significant scene changes are expected, namely:
\begin{itemize}
    \item The camera displacement, with respect to the latest keyframe, exceeds a predefined threshold, both rotation and translation (values in Table \ref{tab:parameters});
    \item Multiple keyframes have been inserted since the last panoptic update.
\end{itemize}

This strategy limits the computational burden of panoptic inference while preserving a class-agnostic object detector.

\subsubsection{Unknown Objects Extraction}
\begin{figure}
    \centering
    \includegraphics[width=0.55\linewidth]{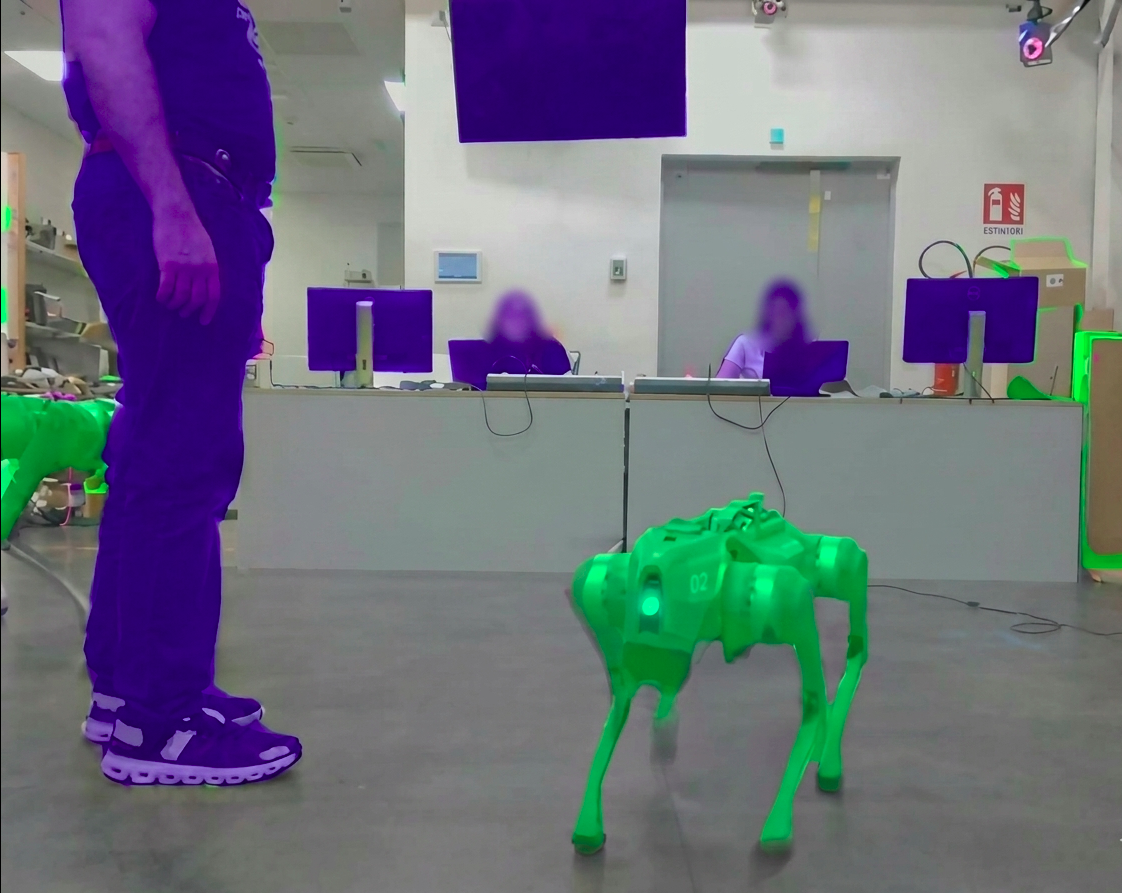}
    \caption{The blue masks represent movable objects with known semantic class. The green masks represent regions containing pixels without a predefined class. Those regions are referred to as \textit{unknown} and may contain one or multiple objects.}
    \label{fig:known_vs_unknown}
\end{figure}

The panoptic segmentation network classifies the image pixels in distinct masked regions, and each mask is labeled as ``Thing", countable object instances, or ``Stuff", uncountable structural entities. Regions not assigned to either category may still contain previously unseen movable objects (see Fig. \ref{fig:known_vs_unknown}). Since segmentation is performed in the image space, a single ``unknown" region can associate pixels that belong to multiple distinct objects or even the background.  
To separate these objects and generate distinct masks, a 3D point-cloud is generated from the unknown region and clustered through the K-means algorithm. The number of clusters K is reduced to a predefined set of values and depends on the depth distribution within the region. The clusters are then projected back into the image space to generate object masks. Spatially disconnected regions are separated into independent object instances.

\subsubsection{Object Association}
The masks labeled as ``Things" and unknown region clusters are added into a unified detected object list. To preserve temporal consistency between consecutive frames, newly detected objects are matched with the one stored in the persistent object list. This matching is performed using the Intersection Over Union (IoU). For each detected mask, the persistent object with the highest IoU score is selected:
\begin{itemize}
    \item If a match is found, the corresponding object state and Kalman filter are updated.
    \item Otherwise, a new object is added to the persistent object database.
\end{itemize}

After the matching step, any object in the persistent object list without a corresponding detection is removed.

\subsection{Propagation Module}
Since running panoptic segmentation on every frame is computationally expensive, object states in intermediate frames are updated through lightweight detection and mask propagation. For each new RGB-D frame, a real-time object detector, YOLO26, is executed to detect known object masks inside the scene. The objects in the persistent object database are associated with the resulting detections, using the Intersection over Union (IoU) metric. For each persistent object, the detected mask with the highest IoU score is selected:
\begin{itemize}
    \item If a match is found, the corresponding object state and Kalman filter are updated.
    \item Otherwise, the stored object is propagated temporally using the proposed mask propagation framework.
\end{itemize}

\subsubsection{Mask Propagation}
This method relies on two main assumptions:
\begin{itemize}
    \item Limited object shape variations across consecutive frames;
    \item Object and camera motion are sufficiently small to preserve spatial continuity; 

\end{itemize}

The prediction step of the object's Kalman filter (KF) is used to estimate the 3D position of the centroid of the object mask, with respect to the world reference frame. The state vector of the KF is represented as
\begin{equation}
    {X}_k =
\begin{bmatrix}
P_x & P_y & P_z & V_x & V_y & V_z
\end{bmatrix}^T
\end{equation}
where $(P_x,P_y,P_z)$ and $(V_x,V_y,V_z)$ denote the object position and velocity, respectively.

Assuming a constant velocity motion model, the predicted state and its covariance are computed as:
\begin{equation}
\hat{{X}}_{k|k-1} = F_k\hat{{X}}_{k-1|k-1}
=
\begin{bmatrix}
{I}_3 & \Delta t\,{I}_3 \\
\mathbf{0} & {I}_3
\end{bmatrix}
\hat{{X}}_{k-1|k-1}
\end{equation}

\begin{equation}
P_{k|k-1} = F_k P_{k-1|k-1} F_k^T + Q_k
\end{equation}

\noindent where $F_k$ is the state transition matrix, ${P}_{k|k-1}$ is the predicted covariance matrix, and ${Q}_k$ is the process noise covariance matrix derived from a white-noise acceleration model with axis-dependent acceleration variances. All state estimates are expressed in the world reference frame.

Once the predicted state $\hat{X}_k^{world}$ is computed, the predicted object centroid in the world reference frame is determined as:
\begin{equation}
    \hat{C}_k ^{world}= [\hat{P}_x, \hat{P}_y, \hat{P}_z]^T
\end{equation}
where
\begin{equation}
    \hat{{X}}_{k|k-1}^{world} = [\hat{P}_x, \hat{P}_y, \hat{P}_z, \hat{V}_x, \hat{V}_y, \hat{V}_z] ^T
\end{equation}

Using the estimation of the camera pose $\hat{T_{WC}}$ (see section \ref{initial_processing}), the predicted mask centroid $\hat{C_k}^{world}$ is transformed into the camera reference frame.
\begin{equation}
    \begin{bmatrix}
        \hat{c}_{k}^{cam} \\
        1
    \end{bmatrix}
    =
    \hat{T}_{CW}
    \begin{bmatrix}
        \hat{C}_{k}^{world} \\
        1
    \end{bmatrix}
\end{equation}
where $\hat{T_{CW}}$ is the transform from world to camera reference frame.

The next step is to compute the 2D predicted mask centroid $\hat{c'_k}$. This is done by projecting the $\hat{c}_k^{cam}$ into the image frame using the camera intrinsic matrix. 
The bounding box associated with the previous mask, $bbox_{k-1}$, stored in the persistent object list, is translated by the 2D displacement between the predicted mask centroid $\hat{c'_k}$ and its position in the previous frame $c'_{k-1}$. 
The shifted bounding box acts as a Region Of Interest (ROI), the area of the current frame where the object is most likely to be. Within this region, the previous mask is evaluated at multiple candidate positions, and the highest scoring candidate is used to temporally propagate the object state. Each candidate is scored through the cost function:

\begin{equation}
    F = S_{D}W_D + S_{C}W_C
    \label{cost_function}
\end{equation}
where $S_{D}$ and $S_{C}$ are the depth and color similarity factors, $W_D$ and $W_C$ are their respective weights.\\

\paragraph{Depth similarity}

The depth similarity factor compares the measured depth of the candidate mask and the predicted depth, from the Kalman filter, projected in the camera reference frame. This score awards the mask position whose centroid is closer to the predicted one:
\begin{equation}
    S_{D} = e^ {-|p_z - \hat{p_z}|}
\end{equation}
where $p_z$ is the depth of the centroid of the candidate mask, and $\hat{p_z}$ is the predicted depth obtained from the Kalman filter prediction step, both expressed in meters.

\paragraph{Color similarity}

The color similarity factor is computed using HSV color histograms. Let $H_{k-1}$ and $H_{cand}$ denote the normalized histograms associated with the previous and candidate masks, respectively. Their similarity is evaluated using the Bhattacharyya distance. Therefore, this factor measures how different the two color distributions are by looking at the overlap between each bin:
\begin{equation}
    d_B(H_{k-1}, H_{cand}) =
    \sqrt{
    1 -
    \sum_{i=1}^{N}
    \sqrt{H_{k-1}(i)\,H_{cand}(i)}
    }
\end{equation}
which is converted into a similarity score as:
\begin{equation}
    S_{C} =1 - d_B(H_{k-1}, H_{cand})
\end{equation}

\noindent where $d_B(H_{prev}, H_{cand})$ is the Bhattacharyya distance, $H_{prev}$ and  $H_{cand}$ are the mask histograms of the previous mask and candidate, and $S_{C}$ is the color similarity factor.

The combination of depth and appearance consistency significantly reduces mask drift toward the background, while preserving robustness to viewpoint and illumination changes.
The candidate position that maximizes the cost function in \eqref{cost_function} is selected as the propagated object mask. Its centroid, projected into the world frame, serves as the Kalman filter measurement $\mathbf z_k=[P_x^m,P_y^m,P_z^m]^T$, with observation matrix $\mathbf H=[\mathbf I_3\;\mathbf 0]$. The noise in the centroid estimate is modeled by a diagonal covariance $R_k\in\mathbb{R}^{3\times3}$ with experimentally tuned variances. After the standard Kalman correction step, the object state in the persistent database is updated accordingly.

After all masks in the persistent object list are propagated, an overlap check is performed, since moving objects may partially occlude background ones. If an overlap is found, the background mask is trimmed to keep only its visible region.

\subsection{Dynamic object filtering}

To determine whether detected movable objects are currently in motion, each tracked object is classified as static or dynamic using a motion-consistency analysis performed on its estimated state. Two criteria are considered.

\subsubsection{Velocity Threshold}
The velocity threshold evaluates the magnitude of the estimated object velocity from the Kalman filter state. An object is classified as dynamic if:
\begin{equation}
    \left\lVert {V_i} \right\rVert_2 > V_{th_i}
\end{equation}
where $V_i$ is the estimated velocity and $V_{th_i}$ is the depth-dependent threshold. $V_{th_i}$ is constrained to a set of predefined and empirically tuned values. A high threshold is assigned to objects in the background, helping to mitigate the effect of depth estimation noise, while a lower threshold is assigned to objects in the near field, preserving sensitivity to the motion of nearby objects. Furthermore, to improve temporal consistency, a hysteresis mechanism is introduced:

\begin{equation}
V_{th,i}=
\begin{cases}
a(p_{z_i}), \quad \text{if } isDyn_{i}^{k-1}=0\\
\alpha a(p_{z_i}), \quad \text{if } isDyn_{i}^{k-1}=1
\end{cases}
\end{equation}
with
\begin{equation}
    a(p{z_i}) = \tau_{low} + \frac{\tau_{high} - \tau_{low}}{1 + e^{- k (p_{z_i} - z_0)}}
\end{equation}
where $p_{z_i}$ is the object $i$ depth, $\tau_{low}$ and $\tau_{high}$ are respectively the sensitivity for near and far objects, $k$ is the transition sharpness parameter, $z_0$ is the characteristic depth and $\alpha <1$ is the hysteresis parameter (values in Table \ref{tab:parameters}).

 \subsubsection{Spatial displacement}
 
 Slow-moving objects may exhibit velocities close to the noise level while undergoing significant long-term motion. Therefore, the spatial displacement criterion evaluates the displacement from the object initial pose.

 \begin{equation}
     D_i = \left\lVert{P_i^k - P_i^0} \right\rVert_2
 \end{equation}
 where $D_i$ is the displacement, and $P_i^k$ and $P_i^0$ are the current and initial position of the object, respectively. An object is classified as dynamic if $D_i > D_{th_i}$, with

\begin{equation}
D_{th,i} =
\begin{cases}
b, \quad \text{if } isDyn_{i}^{k-1}=0,\\
\beta b, \quad \text{if } isDyn_{i}^{k-1}=1,
\end{cases}
\end{equation}

\noindent where $\beta < 1$ is the hysteresis parameter (values in Table \ref{tab:parameters}).

\begin{table}[H]
    \centering
    \setlength{\tabcolsep}{5pt}
    \begin{tabular}{l|c c c}
        \hline
        \textbf{Parameter}&  \textbf{Meaning}& \textbf{Value}\\
         \hline
         $IoU_{th}$&  Intersection Over Union Threshold& 0.3\\
 $\alpha$& Object velocity hysteresis factor&0.8\\
 $k$ & Sharpness transition factor & 8.0 \\
 $z_0$ & Characteristic depth where the transition occurs & 2.75 m \\
 $\tau_{low}$ & Sensitivity for near objects & 0.15 m/s\\
 $\tau_{high}$ & Maximum tolerance for far objects & 0.40 m/s\\
 $\beta$& Object spatial displacement hysteresis factor&0.5\\
 $b$ & Spatial displacement threshold & 0.10 m \\
         $T_n$&  Camera translation threshold& 0.15 m\\
         $R_n$&  Camera rotation threshold& 0.15 rad\\
         \hline
    \end{tabular}
    \caption{Experimentally tuned thresholds}
    \label{tab:parameters}
\end{table}

\subsection{Dynamic features filtering}
Once the mask of each object present in the scene has been classified as dynamic or static, all ORB features whose coordinates lie inside the dynamic masks are discarded. The remaining subset of features is assumed to be from static scene structures or objects and is therefore used for feature matching and camera pose estimation. 

After dynamic filtering, the ORB-SLAM3 pipeline is executed as in the standard configuration. The final camera pose estimation $T_{WC}$ is used for new keyframe insertion by the Tracking Thread, if needed. If a new keyframe is added, the Mapping Thread is executed, and the map is updated with new MapPoints generated by the static features.

Parallel to this, the Loop Closure \& Map Merging Thread runs to refine the map structure and the camera poses if needed.
Consequently, the proposed framework preserves the optimization and relocalization capabilities of ORB-SLAM3 while improving robustness in highly dynamic environments.

\section{RESULTS}
\begin{figure}
    \centering
    \begin{subfigure}{0.49\columnwidth}
        \centering
        \includegraphics[width=\linewidth]{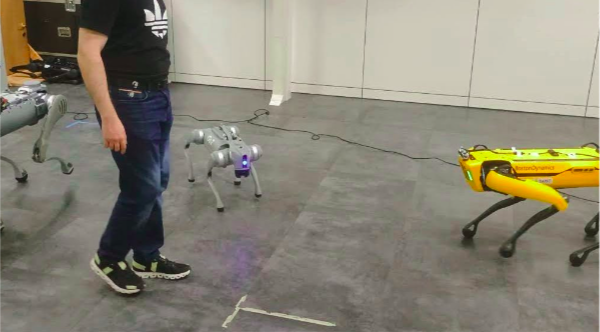}
    \end{subfigure}
    \hfill
    \begin{subfigure}{0.49\columnwidth}
        \centering
        \includegraphics[width=\linewidth]{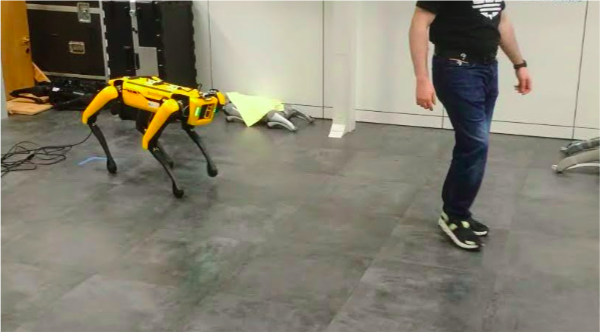}

    \end{subfigure}
    \caption{A scene of go2\_b2\_person\_move (left) and person\_move (right) from an external point of view}
    \label{fig:exp_photo}
\end{figure}

The system was evaluated in terms of camera localization accuracy, measured by the Root Mean Squared Error (RMSE) of the Absolute Pose Error (APE), and computational efficiency,  measured by the frame rate. Two public RGB-D datasets were used, and compared to the ORB-SLAM3 baseline~\cite{orbslam3} and to the dynamic SLAM systems NGD-SLAM~\cite{Zhang2025ngdslam} (semantic-based) and Panoptic-SLAM~\cite{abati2024panopticslam} (class-agnostic). A real-world experiment was conducted to jointly evaluate camera localization and object tracking accuracy (Fig. \ref{fig:exp_photo}). Each experiment was repeated five times and the median result is reported.

\subsection{RGB-D datasets}

The proposed system was tested on sequences from two complementary datasets, to evaluate its robustness under different types of dynamics:
\begin{itemize}
    \item Bonn RGB-D Dynamic Dataset~\cite{8967590} contains highly dynamic scenes with moving people and manipulated objects. In particular the \textit{nonobstructing\_box} sequences show a person moving a little box around, and the \textit{obstructing\_box} sequences show two people moving a wide box that can fully cover the camera view;
    \item Panoptic Dataset~\cite{abati2024panopticslam}, containing quadruped robots and people.
\end{itemize}
These two datasets include objects with different appearance, size and motion characteristics. Moreover, quadruped robots do not belong to any COCO classes used to train the panoptic network, and thus appear as \textit{unknown} moving objects, allowing us to assess the class-agnostic capabilities of our system.

\paragraph{Camera localization accuracy}
Table \ref{tab:all_results} reports localization results on the Bonn RGB-D Dynamic Dataset \cite{8967590} and the Panoptic Dataset \cite{abati2024panopticslam}. Overall, the proposed framework consistently improves camera localization in highly dynamic environments while maintaining competitive performance in scenarios dominated by human motion (\textit{Bonn - Dynamic Sequences}). The largest gains are observed in situations where semantic information alone is insufficient to identify dynamic objects (\textit{Panoptic Dataset}) or where large moving objects produce extensive scene occlusions (\textit{Bonn - Obstructing}).

In \textit{Moving Robots} (Table \ref{tab:all_results}) the scene contains semantically unknown quadruped robots. The proposed method achieves the lowest localization error, reducing the RMSE from 9.7 cm (Panoptic SLAM) and 9.8 cm (NGD-SLAM) to 4.0 cm, a reduction of up to 59\%. This highlights one of the main advantages of using the proposed framework: generalization to previously unseen objects without additional semantic training.

\begin{figure}
    \centering
    \includegraphics[width=\linewidth]{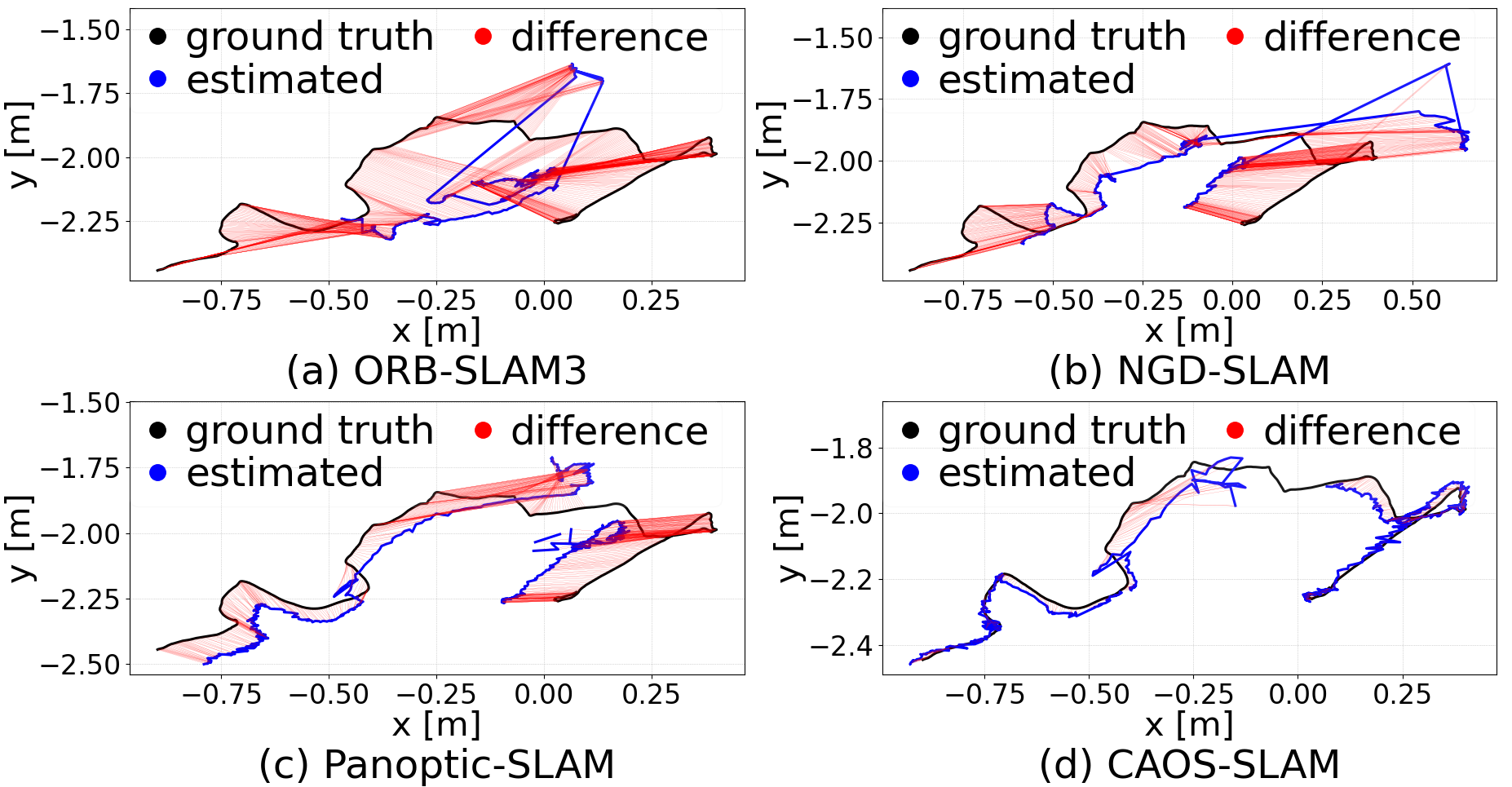}
    \caption{Comparison between the ground truth and the estimated trajectories on the \textit{moving obstructing box 2} sequence}
    \label{fig:est_gt_moving_obs}
\end{figure}

The largest gains occur in the \textit{obstructing box} sequences, where a large moving box repeatedly occludes a substantial part of the scene, sharply reducing observable static features. By filtering all features on the dynamic box, which semantic-based approaches fail to detect, the proposed method mitigates map corruption and localization drift, enabling later relocalization and loop closure once sufficient static structure reappears (Fig. \ref{fig:est_gt_moving_obs}).

A different behavior is observed in human-centered sequences, such as \textit{crowd} (Table \ref{tab:all_results}), where semantic-based methods like NGD-SLAM achieve comparable or slightly better results, as expected, since people belong to the COCO classes they are trained on.
Nevertheless, the proposed method remains aligned with the state of the art and consistently outperforms ORB-SLAM3.
The main contribution of our hybrid system lies in extending semantic dynamic SLAM to environments with unknown dynamic objects or severe dynamic occlusions, where semantic segmentation alone is insufficient.

\begin{table}[t]
    \centering
    \scriptsize
    \setlength{\tabcolsep}{5pt}
    \caption{Performance comparison of the RMSE APE [m]. The best result is in bold, and the second best is underlined.}
    \label{tab:all_results}
    \begin{tabular}{l|ccc|cc}
    \hline
\textbf{Sequence} &
    \makecell{\textbf{ORB-}\\ \textbf{SLAM3}} &
    \makecell{\textbf{Panoptic-}\\ \textbf{SLAM}} &
    \makecell{\textbf{NGD-}\\ \textbf{SLAM}} &
    \makecell{\textbf{Only}\\ \textbf{Panoptic}} &
    \makecell{\textbf{CAOS-SLAM}\\ \textbf{(Ours)}} \\
    \hline

        \multicolumn{6}{c}{\textbf{Bonn - Non-Obstructing Box}} \\
        \hline
        Moving    & 0.3775 & 0.0273 & \textbf{0.0211} & 0.1271 & \underline{0.0272} \\
        Moving 2   & 0.0411 & \textbf{0.0329} & 0.0395 & 0.0423 & \underline{0.0345} \\
        Placing  & 0.8513 & 0.0686 & 0.3479 & \textbf{0.0405} & \textbf{0.0405} \\
        Placing 2  & \underline{0.0220} & 0.0235 & \textbf{0.0181} & 0.0262 & 0.0241 \\
        Placing 3  & 0.2402 & \textbf{0.0299} & 0.0354 & \underline{0.0338} & 0.0349 \\
        Removing  & \underline{0.0166} & 0.0179 & 0.0171 & 0.0187 & \textbf{0.0165} \\
        Removing 2 & \textbf{0.0199} & 0.0290 & \underline{0.0202} &0.0229 & 0.0228 \\
        \hline
        
        \multicolumn{6}{c}{\textbf{Bonn - Dynamic Sequences}} \\
        \hline
        Person   & 0.5885 & \textbf{0.0442} & 0.0524 & \underline{0.0481}& 0.0515 \\
        Person 2  & 0.7727 & 0.0543 & 0.0581 & \underline{0.0532} & \textbf{0.0530} \\
        Crowd1            & 0.4979 & 0.1107 & \underline{0.0181} & 0.0375 & \textbf{0.0156} \\
        Crowd2            & 0.5707 & 0.4153 & \textbf{0.0222}& \underline{0.0234} & 0.0293 \\
        Crowd3            & 0.2070 & 0.0306 & 0.0306 & \textbf{0.0290} & \underline{0.0305} \\
        Synchronous       & 0.8080 & 0.1897 & \textbf{0.0229} & 0.0417 & \underline{0.0335} \\
        Synchronous 2 & 1.1780 & \underline{0.0089} & 0.0097 & \textbf{0.0078} & \underline{0.0089} \\
        \hline
        
        \multicolumn{6}{c}{\textbf{Panoptic Dataset}} \\
        \hline
        Baseline          & 0.0645 & \textbf{0.0570} & 0.0595 & 0.0595& \underline{0.0594} \\
        Moving Robots    & 0.3808 & 0.0970 & \underline{0.0983} & 0.0986& \textbf{0.0400} \\
        Robots people    & 0.2723 & \underline{0.0557} & 0.0571 & 0.0603 &\textbf{0.0517} \\
        \hline
        
        \multicolumn{6}{c}{\textbf{Bonn - Obstructing Box} } \\
        \hline
        Moving    & 0.3676 & 0.2583 & 0.3678 & \underline{0.1867} & \textbf{0.1701} \\
        Moving 2   & 0.8980 & 0.2304 & 0.4897 & \underline{0.1364}& \textbf{0.0274} \\
        Placing   & 0.2992 & 0.2398 & 0.3155 & \textbf{0.1880} & \underline{0.2276} \\
        Removing  & 0.4810 & \underline{0.2548} & 0.4315 & 0.2788& \textbf{0.2543} \\
        \hline
        
    \end{tabular}
\end{table}

\paragraph{Computational performance \& ablation study}
\begin{table}[ht]
\centering
\small
\resizebox{\linewidth}{!}{%
\begin{tabular}{l|ccc|cc}
\hline
\textbf{} &
    \makecell{\textbf{ORB-}\\ \textbf{SLAM3}} &
    \makecell{\textbf{Panoptic-}\\ \textbf{SLAM}} &
    \makecell{\textbf{NGD-}\\ \textbf{SLAM}} &
    \makecell{\textbf{Only}\\ \textbf{Panoptic}} &
    \makecell{\textbf{CAOS-SLAM}\\ \textbf{(Ours)}} \\
    \hline
Median FPS & 49.33 & 2.618 & 82.64 & 2.207 & 30.61 \\
Mean FPS   & 50.35 & 2.583 & 97.72 & 2.189 & 24.73 \\
\hline
\end{tabular}%
}
\caption{Time performance comparison in frames per second (FPS).}
\label{tab:time_table}
\end{table}
The runtime performance of the evaluated methods is summarized in Table \ref{tab:time_table}. The tests were conducted on a laptop equipped with an NVIDIA GeForce RTX 3050 with 4GB of VRAM and an Intel Core i7-12700H CPU, using YOLO26-seg nano for lightweight object detection and KMaX-DeepLab R50~\cite{10.1007/978-3-031-19818-2_17} for panoptic segmentation.

Table \ref{tab:time_table} compares CAOS-SLAM with ORB-SLAM3, Panoptic-SLAM, NGD-SLAM and a system variant running the panoptic model each frame, \textit{Only Panoptic}. The proposed framework achieves a median speed of about 30 FPS, retaining the benefits of panoptic segmentation in real time. Although heavier than ORB-SLAM3 due to the added perception module, it remains suitable for online robotic applications.

Disabling mask propagation (\textit{Only Panoptic}) reduces the median speed from 30.6 FPS to just 2.2 FPS, without a consistent accuracy gain (Table \ref{tab:all_results}), as the full system generally achieves comparable or better pose accuracy at a fraction of the computational cost. This confirms that decoupling object discovery from tracking is essential for real-time performance without sacrificing the accuracy of object-aware SLAM.

\subsection{Parameter Sensitivity Analysis}

\begin{figure}
    \centering
    \includegraphics[width=\linewidth]{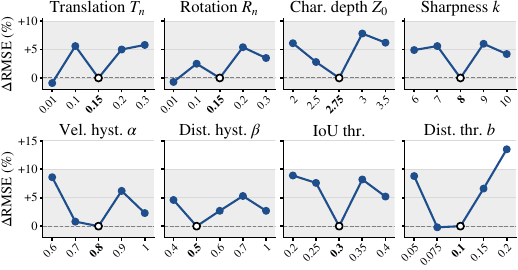}
    \caption{Parameter Sensitivity graph. Per each hyperparameter, the median percentage deviation in the RMSE is plotted. The white dots represent the default values}
    \label{fig:threshold_sweep}
\end{figure}

A sensitivity analysis was conducted to evaluate the robustness of the proposed method to its key parameters, whose default values are listed in Table~\ref{tab:parameters}. Five values were independently evaluated for each parameter on six dynamic sequences, keeping all others fixed. Fig. \ref{fig:threshold_sweep} shows the median percentage deviation in ATE-RMSE from the baseline configuration.
Overall, the median deviation from baseline remains below 9\%, with $T_n$ and $R_n$ being the least sensitive, with 5.5 \% and 3.0 \%, respectively. Only the spatial displacement threshold $b$, at its most permissive tested values, produces a larger median deviation (13\%).
These results indicate that the system's overall localization accuracy is largely stable with respect to these parameters.
\subsection{Experiments}
\begin{figure}[t]
    \centering

    \begin{subfigure}{0.5\columnwidth}
        \centering
        \includegraphics[width=\linewidth, height=4.3cm]{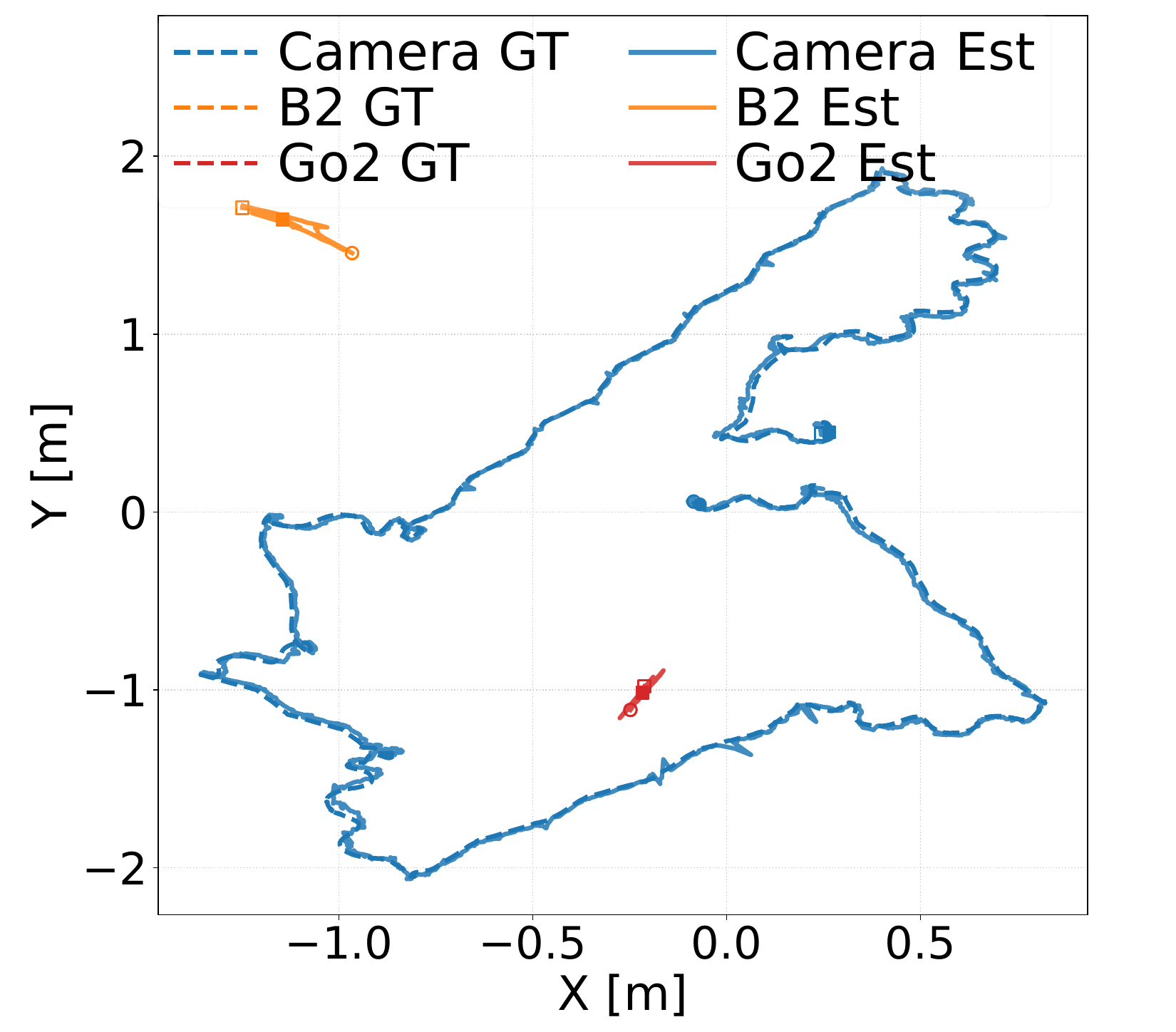}
        \caption{Baseline}\label{baseline}
    \end{subfigure}
    \hfill
    \begin{subfigure}{0.48\columnwidth}
        \centering
        \includegraphics[width=\linewidth, height=4.3cm]{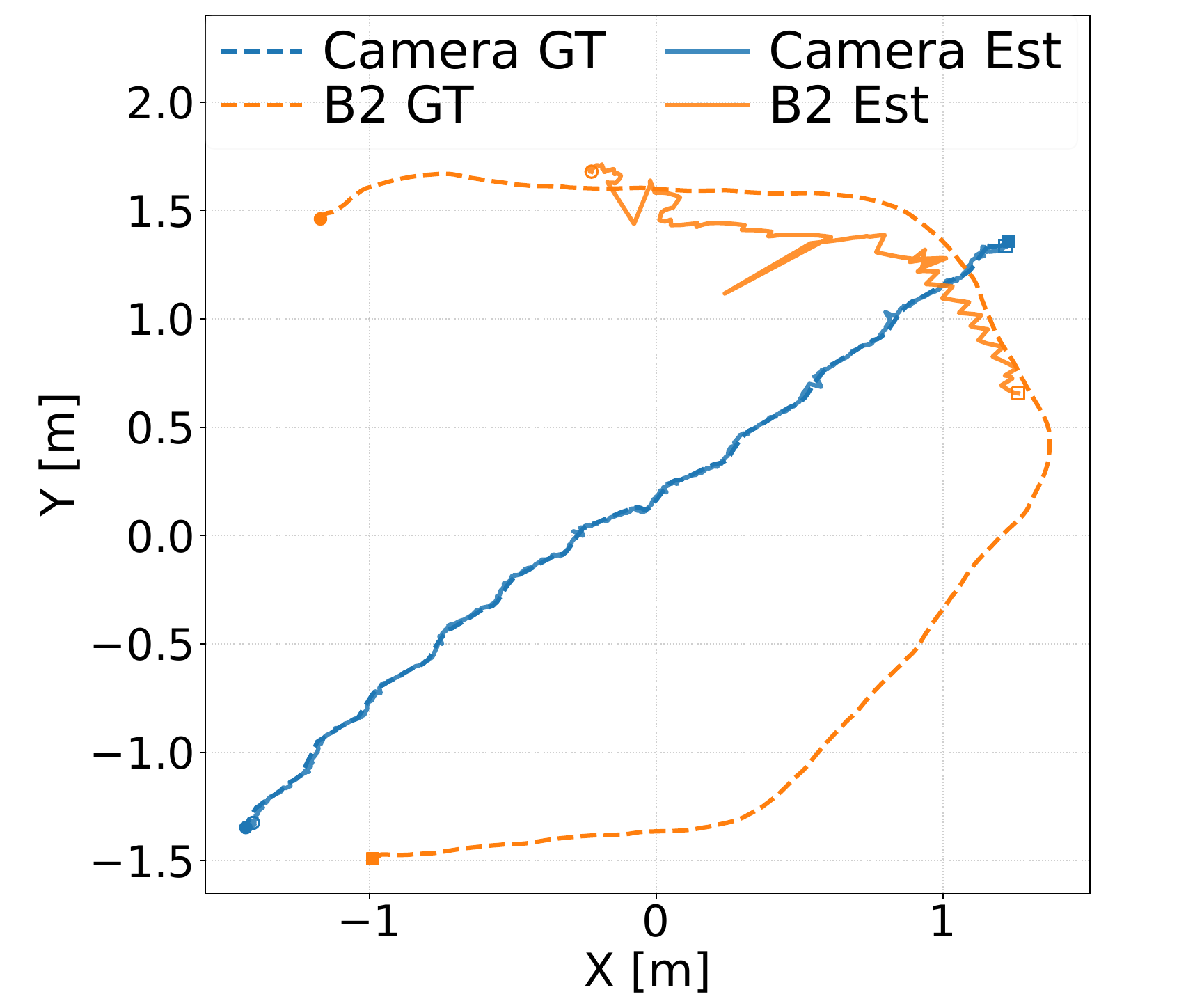}
        \caption{B2 move}
    \end{subfigure}

    \vspace{2mm}

    \begin{subfigure}{0.48\columnwidth}
        \centering
        \includegraphics[width=\linewidth, height=4.3cm]{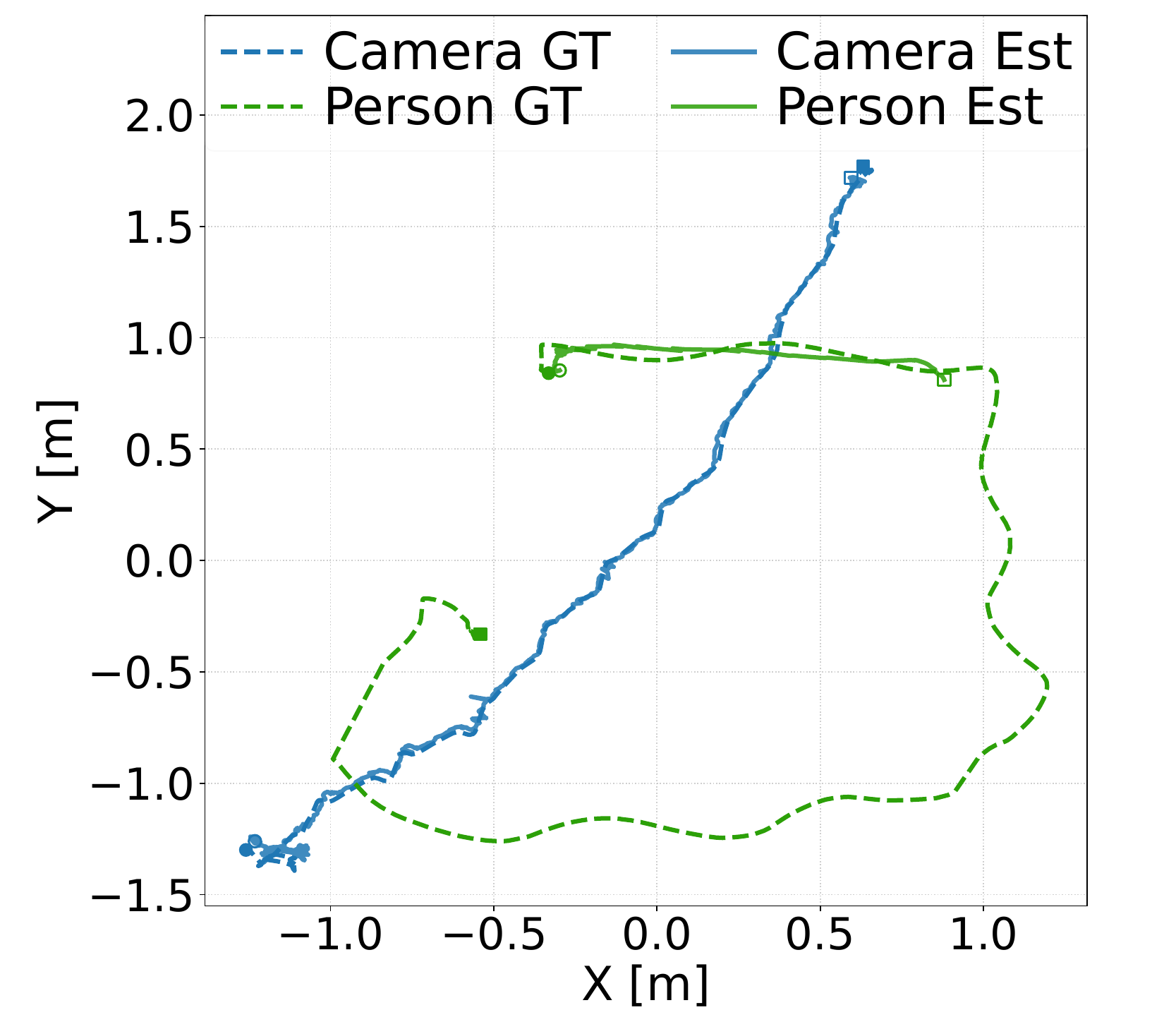}
        \caption{Person Move}
    \end{subfigure}
    \begin{subfigure}{0.5\columnwidth}
        \centering
        \includegraphics[width=\linewidth, height=4.3cm]{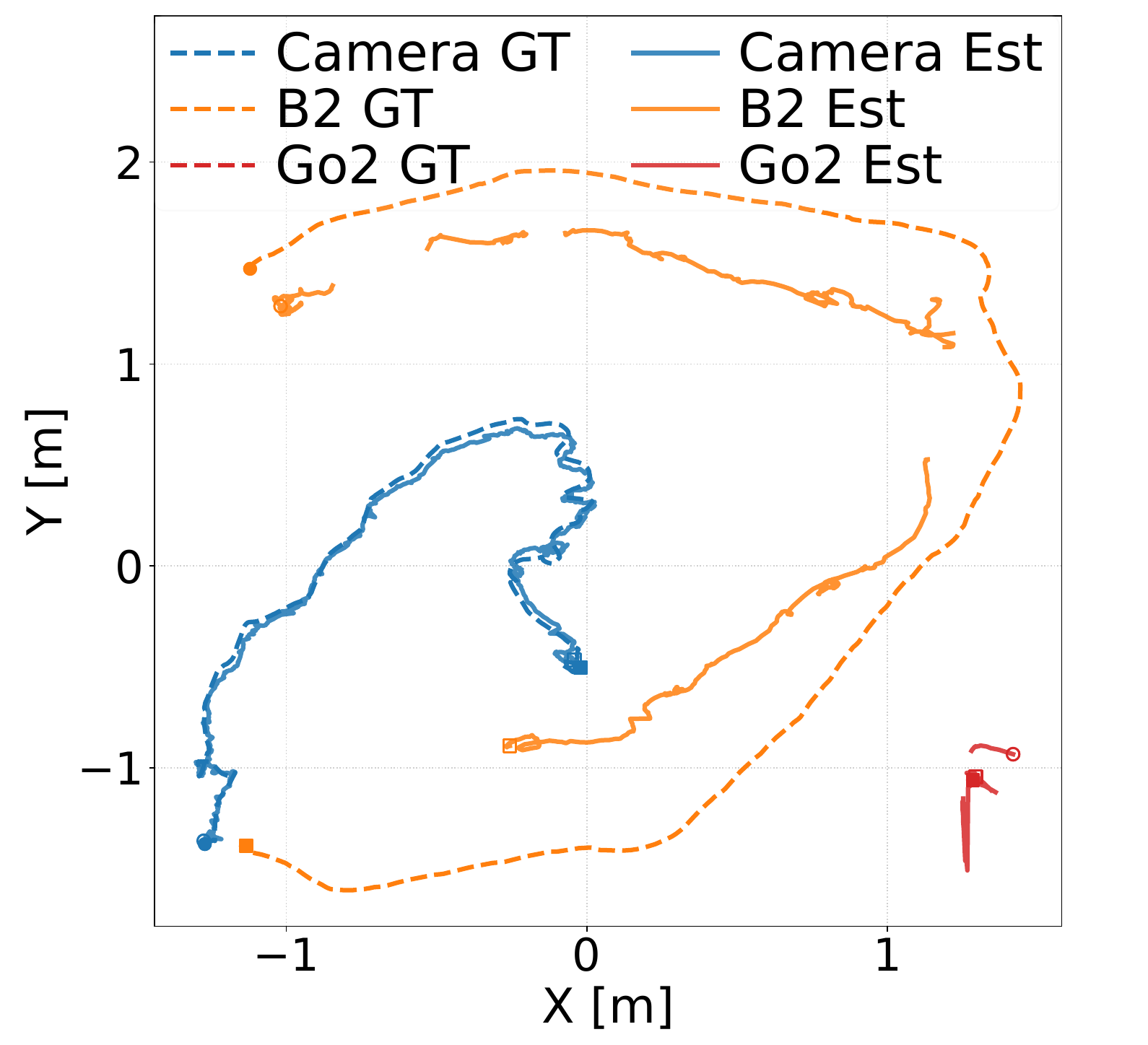}
        \caption{B2 follow}
        \label{fig:b2_follow}
    \end{subfigure}

    \caption{Estimated camera and object trajectories compared to ground truth}
    \label{fig:est_gt_exp}
\end{figure}

The real-world experiments jointly evaluate camera pose estimation and object tracking accuracy, while illustrating a potential application of the proposed system. Ground-truth trajectories of the camera and moving entities  were recorded with a Vicon Motion Capture system.
A RealSense D435 RGB-D camera was mounted on a Boston Dynamics SPOT robot, while a Unitree B2 robot, a Unitree Go2 robot, and a person acted as dynamic agents. Eight sequences were recorded, covering static, single-agent and multi-agent scenarios with varying motion and scene complexity. In \textit{baseline}, the B2 and Go2 were present but static. In \textit{B2 move} and \textit{B2 follow} the B2 moved along a semicircle while SPOT either walked straight or followed it. \textit{person move} and \textit{person follow} used the same camera motions with a person walking a semicircle, while in \textit{person random} the person moved randomly. Finally, in \textit{B2 Go2 move} the two robots moved while SPOT moved back and forth toward them, and in \textit{B2 Go2 person move}, the same motions were used with also the person involved. Fig.~\ref{fig:est_gt_exp} compares the estimated and ground-truth camera and object trajectories for four representative sequences.

\paragraph{Camera localization accuracy}

The camera trajectory results in Table \ref{tab:cam_RMSE} confirm the trends observed on the public datasets. 
The experiments highlight that the accuracy in the camera trajectory varies depending on the scene complexity and motion characteristics. The most significant improvement is obtained in the \textit{B2 follow} sequence (Fig.~\ref{fig:b2_follow}), where the camera is following the Unitree B2 robot. The localization error is reduced by approximately 74\% compared to NGD-SLAM. 
In contrast, the performance gap is reduced in sequences involving human motion. As observed in the public datasets, semantic segmentation provides reliable identification of the dominant dynamic objects in scenarios involving mainly semantically known entities.
Overall, the real-world experiments confirm that the proposed hybrid architecture is particularly effective for robotic applications in environments containing heterogeneous dynamic objects, while maintaining real-time performance.

\paragraph{Object localization accuracy}

Table~\ref{tab:objects_RMSE} reports the RMSE APE of the estimated object trajectories. Since object poses are estimated starting from the camera pose, the results show a correlation between the object trajectory and camera localization accuracy. The two sequences with the highest camera error, \textit{B2 follow} and \textit{B2 Go2 person move}, also show some of the highest object tracking errors.

Since the mask centroid used for tracking depends on the visible surface of the objects, object size plays an important role. The B2 robot, the largest articulated robot in the experiments, consistently presents higher localization errors compared to the smaller Go2, even when the camera is accurately localized. In \textit{baseline} (Fig.~\ref{baseline}), for instance, the B2 error is 0.24~m against 0.08~m  for Go2, with a camera error of only 0.02~m. This systematic localization bias can be seen in Fig.~\ref{fig:b2_follow}.

\begin{table}[ht]
\centering
\setlength{\tabcolsep}{5pt}
\caption{RMSE APE [m] comparison on the camera tracking error. The best result is in bold, and the second best is underlined.}
\label{tab:cam_RMSE}
\begin{tabular}{l|ccc|c}
\hline
\textbf{Sequences} &
    \makecell{\textbf{ORB-}\\ \textbf{SLAM3}} &
    \makecell{\textbf{Panoptic-}\\ \textbf{SLAM}} &
    \makecell{\textbf{NGD-}\\ \textbf{SLAM}} &
    \makecell{\textbf{CAOS-SLAM}\\ \textbf{(Ours)}} \\
    \hline
baseline        & \textbf{0.0182}&0.0232& 0.0296& \underline{0.0198}\\
B2 move   & 0.0425& \textbf{0.0177}& 0.0224& \underline{0.0189}\\
B2 follow   & 0.7330& 0.7236& \underline{0.6779}& \textbf{0.1754}\\
 person move& 0.3495& - &\underline{0.0915}&\textbf{0.0751}\\
 person random& 0.3824& \underline{0.0235}&\textbf{0.0210}&0.0317\\
 person follow& 0.6321& 0.4154& \underline{0.2068}&\textbf{0.0375}\\
 B2 Go2 move& 0.5931& - & \textbf{0.0216}& \underline{0.0502}\\
 B2 Go2 person move& 0.6313& - & \underline{0.2481}&\textbf{0.1970}\\
 \hline
\end{tabular}
\end{table}

\begin{table}[ht]
\centering
\caption{RMSE APE [m] object tracking error}
\label{tab:objects_RMSE}
\begin{tabular}{l|ccc}
\hline
\textbf{Sequence} & B2& Go2& Person\\
\hline
baseline        & 0.2383& 0.0762& -\\
B2 move   & 0.2315& -& -\\
B2 follow   & 0.3799& 0.1367& -\\
 person move& -& -&0.0652\\
 person random& -& -& 0.1702\\
 person follow& -& -& 0.2644\\
 B2 Go2 move& 0.2137& 0.1128& -\\
 B2 Go2 person move& 0.2478& 0.2959& 0.2886\\
 \hline
\end{tabular}
\end{table}

\subsection{Limitations}
Object tracking accuracy is bounded by camera localization accuracy. Since object poses are computed starting from the projection of image observations into the world frame using the estimated camera pose, errors in camera localization directly affect the object positions. Moreover, corrections from loop closure and global map optimization are not retroactively applied to previously estimated object positions, so these errors remain in the reconstructed object trajectories.

As mentioned above, centroid-based object representation, while computationally efficient, introduces localization biases for large and geometrically complex objects. Since the mask centroid depends on the visible surface of the object, changes in viewpoint or occlusions can produce centroid displacements even when the physical object position remains unchanged.

Another limitation concerns object ID persistence. The proposed implementation maintains consistent object IDs while the object remains visible in the scene. However, when this visual continuity is interrupted, the object is occluded or it exits the frame, the system does not perform object re-identification. Consequently, if the object reappears in the scene it may be assigned with a new ID. Although this limitation does not directly affect the camera localization accuracy, it reduces the temporal consistency of long-term object tracking.

\section{CONCLUSIONS}
This paper presented CAOS-SLAM, a dynamic-aware vSLAM framework, designed to improve camera localization robustness in highly dynamic environments while preserving real-time performance and tracking of the dynamic objects in the scene. To balance segmentation quality and computational efficiency, the framework combines panoptic segmentation on selected frames with lightweight temporal mask propagation and Kalman-filter-based object tracking, integrated into ORB-SLAM3.

Experimental evaluation on the Bonn RGB-D Dynamic Dataset, the Panoptic Dataset and real-world experiments demonstrated that the proposed approach improves camera localization accuracy in challenging dynamic scenarios, while maintaining real-time operation. The biggest improvements were observed in scenarios containing large moving obstacles, severe scene occlusions, and dynamic objects not represented by predefined semantic classes, highlighting the advantages of the proposed class-agnostic framework.

Despite these promising results, some limitations remain. Future work will address the centroid-based tracking bias and the lack of object re-identification through richer object representations and adaptive motion models.

\bibliographystyle{IEEEtran}
\bibliography{main}

\end{document}